\documentclass[11pt,letterpaper]{article}
\usepackage{emnlp2017}
\usepackage{times}
\usepackage{latexsym}

\usepackage{graphicx}
\usepackage{booktabs}
\usepackage{tabularx}
\usepackage{amsmath}

\usepackage{color}

\emnlpfinalcopy

\def\emnlppaperid{***}

\title{Auxiliary Objectives for Neural Error Detection Models}

\author{\hspace{-0.5cm}Marek Rei\\
	    \hspace{-0.5cm}The ALTA Institute\\
	    \hspace{-0.5cm}Computer Laboratory\\
	    \hspace{-0.5cm}University of Cambridge\\
        \hspace{-0.5cm}United Kingdom\\
	    \hspace{-0.5cm}{\tt marek.rei@cl.cam.ac.uk}
	    \And
	    \hspace{0.5cm}Helen Yannakoudakis\\
	    \hspace{0.5cm}The ALTA Institute\\
	    \hspace{0.5cm}Computer Laboratory\\
	    \hspace{0.5cm}University of Cambridge\\
        \hspace{0.5cm}United Kingdom\\
        \hspace{0.5cm}{\tt helen.yannakoudakis@cl.cam.ac.uk}}

\date{}

\begin{document}

\maketitle

\begin{abstract}
We investigate the utility of different auxiliary objectives and training strategies within a neural sequence labeling approach to error detection in learner writing. 
Auxiliary costs provide the model with additional linguistic information, allowing it to learn general-purpose compositional features that can then be exploited for other objectives.
Our experiments show that a joint learning approach trained with parallel labels on in-domain data improves performance over the previous best error detection system. 
While the resulting model has the same number of parameters, the additional objectives allow it to be optimised more efficiently and achieve better performance.
\end{abstract}

\section{Introduction}


Automatic error detection systems for learner writing need to identify various types of error in text, ranging from incorrect uses of function words, such articles and prepositions, to semantic anomalies in content words, such as adjective--noun combinations.
To tackle the scarcity of error-annotated training data, previous work has investigated the utility of automatically generated ungrammatical data \cite{foster2009generrate,felice2014generating}, as well as explored learning from native well-formed data \cite{rozovskaya2016grammatical,gamon2010using}. 

In this work, we investigate the utility of supplementing error detection frameworks with additional linguistic information that can be extracted from the available error-annotated learner data. 
We construct a neural sequence labeling system for error detection that allows us to learn better representations of language composition and detect errors in context more accurately. 
In addition to predicting the binary error labels, we experiment with also predicting additional information for each token, including token frequency and the specific error type,
which can be extracted from the existing data, as well as part-of-speech (POS) tags and dependency relations, which can be generated automatically using readily available toolkits. 

These auxiliary objectives provide the sequence labeling model with additional linguistic information, allowing it to learn useful compositional features that can then be exploited for error detection. This can be seen as a type of multi-task learning, where the model learns better compositional features via shared representations with related tasks. 
While common approaches to multi-task learning require randomly switching between different tasks and datasets, we demonstrate that a joint learning approach trained on in-domain data with parallel labels substantially improves error detection performance on two different datasets. 
In addition, the auxiliary labels are only required during the training process, resulting in a better model with the same number of parameters. 

In the following sections, we describe our approach to the task, systematically compare the informativeness of various auxiliary loss functions, investigate alternative training strategies, and examine the effect of additional training data. 

\section{Error Detection Model}
\label{sec:errordet}

In addition to the scarcity of errors in the training data (i.e., the majority of tokens are correct), recent research has highlighted the variability in manual correction of
writing errors: re-annotation of the CoNLL 2014 shared task test set
by 10 annotators demonstrated that even humans have great difficulty in agreeing how to correct writing errors \cite{bryantfar}. Given the challenges of the all-errors correction task, previous research has demonstrated that detection models can \textit{detect} more errors than systems focusing on correction \cite{Rei2016}, and therefore provide more extensive feedback to the learner.  

Following \newcite{Rei2016}, we treat error detection as a sequence labeling task -- each token in the input sentence is assigned a label, indicating whether it is correct or incorrect given the current context -- and construct a bidirectional recurrent neural network for detecting writing errors.
The model is given a sequence of tokens as input, which are then mapped to a sequence of distributed word embeddings $[x_1, ..., x_T]$.
These embeddings are then given as input to a bidirectional LSTM \cite{Hochreiter1997} moving through the sentence in both directions. At each step, the LSTM calculates a new hidden representation based on the current token embedding and the hidden state from the previous step. 

\begin{equation}
h_t^{(f)} = \textrm{LSTM}(x_t, h_{t-1}^{(f)})
\end{equation}
\begin{equation}
h_t^{(b)} = \textrm{LSTM}(x_t, h_{t+1}^{(b)})
\end{equation}

Next, the network includes a {tanh}-activated feedforward layer, using the hidden states from both LSTMs as input, allowing the model to learn more complex higher-level features.
By combining the hidden states from both directions, we are able to have a vector that represents a specific token but also takes into account context on both sides:
\begin{equation}
d_t = \textrm{tanh}(W_f h_t^{(f)} + W_b h_t^{(b)})
\end{equation}

\noindent where $W_f$ and $W_b$ are fully-connected weight matrices.

The final layer calculates label predictions based on the layer $d_t$.
The softmax activation function is used to output a normalised probability distribution over all the possible labels for each token:
\begin{equation}
y_t = \textrm{softmax}(W_y d_t)
\end{equation}

\noindent where $W_y$ is a weight matrix and $y_t$ is a vector with a position for each possible label. In order to find the predicted label, we return the element with the highest predicted value.

The model is optimised using cross entropy, which is equivalent to optimising the negative log-likelihood of the correct labels:

\begin{equation}
E = - \sum_t \sum_k \widetilde{y}_{t,k} \, \textrm{log}(y_{t,k})
\end{equation}

\noindent where $y_{t,k}$ is the predicted probability of token $t$ having label $k$, and $\widetilde{y}_{t,k}$ has the value $1$ if the correct label for token $t$ is $k$, and the value $0$ otherwise.

We also make use of the character-level extension described by \newcite{Rei2016a}. Each token is separated into individual characters and mapped to character embeddings. Using a bidirectional LSTM and a hidden feedforward component, the character vectors are composed into a character-based token representation. Finally, a dynamic gating function is used to combine this representation with a regular token embedding, taking advantage of both approaches. This component allows the model to capture useful morphological and character-based patterns, in addition to learning individual token-level vectors of common tokens.

\begin{table*}[t]
\small
\begin{tabular}{rcccccccccccc} \toprule
\textbf{words} & My & husband & was & following & a & course & all & the & week & in & Berne & .\\
\textbf{target} & c & c & c & i & c & c & c & i & c & c & c & c\\
\textbf{freq} & 5 & 3 & 8 & 4 & 8 & 5 & 7 & 9 & 5 & 8 & 0 & 10\\
\textbf{lang} & fr & fr & fr & fr & fr & fr & fr & fr & fr & fr & fr & fr\\
\textbf{error} & c & c & c & RV & c & c & c & UD & c & c & c & c\\
\textbf{POS} & APP\$ & NN1 & VBDZ & VVG & AT1 & NN1 & DB & AT & NNT1 & II & NP1 & .\\
\textbf{GR} & det & ncsubj & aux & null & det & dobj & ncmod & det & ncmod & ncmod & dobj & null\\
\bottomrule
\end{tabular}
\caption{Alternative labels for an example sentence from the FCE training data.}
\label{tab:example}
\end{table*}

\section{Auxiliary Loss Functions}
\label{sec:auxloss}

The model in Section \ref{sec:errordet} learns to assign error labels to tokens based on the manual annotation available in the training data. 
However, there are nearly limitless ways of making writing errors and learning them all explicitly from hand-annotated examples is not feasible. 
In addition, writing errors can be very sparse, leaving the system with very little useful training data for learning error patterns. 
In order to train models that generalise well with limited training examples, we would want to encourage them to learn more generic patterns of language, grammar, syntax and composition, which can then be exploited for error detection.  

Multi-task learning allows models to learn from multiple objectives via shared representations, using information from related tasks to boost performance on tasks for which there is limited target data. For example, \newcite{Plank2016} explored the option of using word frequency as an auxiliary loss function for part-of-speech (POS) tagging. 
\newcite{Rei2017} describe a semi-supervised framework for multi-task learning, integrating language modeling as an additional objective.
Following this work, we adapt auxiliary objectives for the task of error detection, and further experiment with a larger set of possible objectives.
Instead of only predicting the correctness of each token in context, we extend the system to predict additional information and labels for every token. The information from these auxiliary objectives is propagated into the weights of the model during training, without requiring the extra labels at testing time.
While common neural approaches to multi-task learning switch randomly between different tasks and datasets, we use a joint learning approach trained on in-domain data only.

The lower parts of the model function similarly to the system described in Section \ref{sec:errordet}. Token representations are first passed through a bidirectional LSTM in order to build context-specific representations. After that, each separate objective is assigned an individual hidden layer:

\begin{equation}
d_t^{(n)} = W_f^{(n)} h_t^{(f)} + W_b^{(n)} h_t^{(b)}
\end{equation}

\noindent where $ W_f^{(n)}$ and $W_b^{(n)}$ are weight matrices specific to the $n$-th task. While the recurrent components are shared between all objectives, the hidden layers allow parts of the model to be customised for a specific task, learning higher-level features and controlling how the information from forward- and backward-moving LSTMs is combined.

\begin{figure}[t]
     \centering 
	\includegraphics[width=0.8\linewidth]{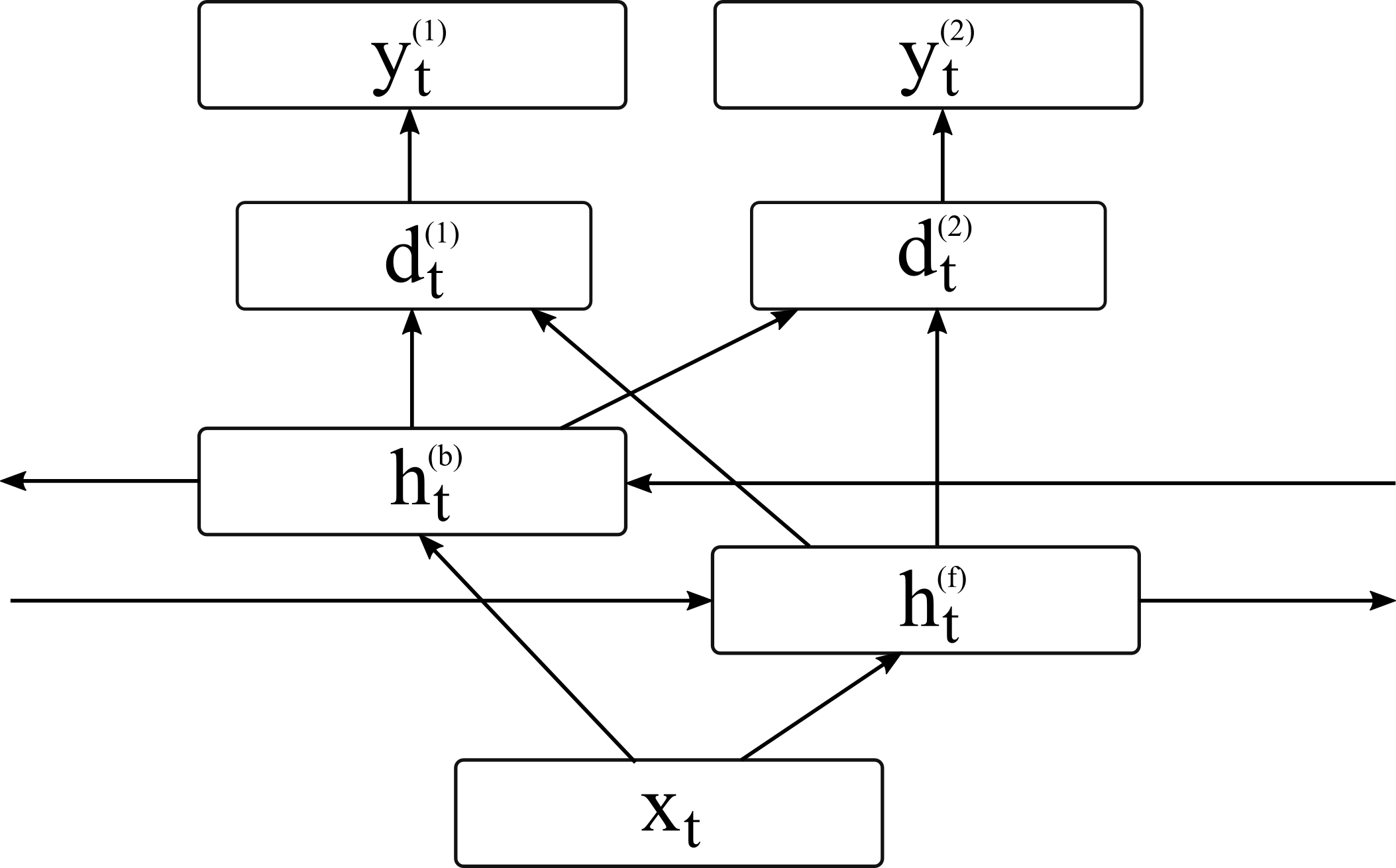}
	\caption{The bidirectional recurrent architecture for one time-step, using one main objective and one auxiliary objective.}
	\label{fig:multitask-network}
\end{figure}

Next, a task-specific output distribution is calculated based on $d_t^{(n)}$:

\begin{equation}
y_t^{(n)} = \textrm{softmax}(W_y^{(n)} d_t^{(n)})
\end{equation}

\noindent where $W_y^{(n)}$ is a weight matrix and $y_t^{(n)}$ has the dimensionality of the total number of labels for the $n$-th task. Figure \ref{fig:multitask-network} presents a diagram of the network with $n=2$, although the number of possible auxiliary tasks can also be larger.

The whole model is optimised by minimising the cross-entropy for every task and every token:

\begin{equation}
E = - \sum_t \sum_n \sum_k \alpha_n \cdot \widetilde{y}_{t,k}^{(n)} \cdot \textrm{log}(y_{t,k}^{(n)})
\end{equation}

\noindent where $y_{t,k}^{(n)}$ is the predicted probability of the $t$-th token having label $k$ for the $n$-th task; $\widetilde{y}_{t,k}^{(n)}$ has value $1$ only if that label is correct, and $0$ otherwise; $\alpha_n$ is the weight for task $n$. Since our main goal is to develop more accurate error detection models, $\alpha_n$ allows us to control how much the model depends on the $n$-th auxiliary task. For example, setting the value of $\alpha_n$ to $0.1$ means any updates for the $n$-th task will have 10 times less importance. We tune a specific weight for each task by trying values $[0.05, 0.1, 0.2, 0.5, 1.0]$ and choosing the ones that achieved the highest result on the development data.

\begin{table*}[t]
\centering
\scalebox{0.9}{
\begin{tabular}{l|ccccc|ccccc} \toprule
 & \multicolumn{5}{c|}{\small{FCE DEV}} & \multicolumn{5}{c}{\small{FCE TEST}}\\
 & predicted & correct & P & R & $F_{0.5}$ & predicted & correct & P & R & $F_{0.5}$\\\midrule
R\&Y (2016) & - & - & 54.5 & 28.2 & 46.0 & 3898 & 1798 & 46.1 & 28.5 & 41.1\\\midrule
Main system & 1837 & 1140 & 62.3 & 24.6 & 47.6 & 2653 & 1468 & 55.7 & 23.3 & 43.4\\
+ frequency & 1870 & 1111 & 59.7 & 23.9 & 45.8 & 2702 & 1461 & 54.4 & 23.2 & 42.7\\
+ language & 1929 & 1150 & 60.4 & 24.8 & 46.6 & 2690 & 1458 & 54.9 & 23.1 & 42.8\\
+ errors & 1905 & 1206 & \textbf{63.3} & 26.0 & 49.2 & 2778 & 1584 & 57.0 & 25.1 & 45.5\\
+ POS & 2199 & 1334 & 60.7 & \textbf{28.8} & 49.7 & 3322 & 1803 & 54.3 & \textbf{28.6} & 46.0\\
+ GR & 1952 & 1207 & 62.1 & 26.0 & 48.4 & 2887 & 1654 & \textbf{57.9} & 26.2 &46.4\\
+ err POS GR & 2087 & 1320 & 63.2 & 28.4 & \textbf{50.8} & 3090 & 1781.0 & 57.7 & 28.3 & \textbf{47.7}\\\bottomrule
\end{tabular}
}
\caption{Error detection results on the FCE dataset using different auxiliary loss functions.}
\label{tab:resultsfce}
\end{table*}

The main goal of our system is to classify tokens as being correct or incorrect, and this objective is included in all configurations. 
In addition, we experiment with a number of auxiliary loss objectives that are only required during training:

\begin{itemize}
\item \textbf{frequency}: \newcite{Plank2016} propose using word frequency as an additional objective for POS tagging, since words with certain POS tags can be more likely to belong to specific frequency groups. The frequency of a token $w$ in the training corpus is discretized as $\textrm{int}(\textrm{log}(\textrm{freq}_{\textrm{train}}(w))$ and used as an auxiliary label.
\item \textbf{error type}: While the task is defined as binary classification, available learner data also contains more fine-grained labels per error. For example, the FCE \cite{Yannakoudakis2011} training set has 75 different labels for individual error types, such as missing determiners or incorrect verb forms. By giving the model access to these labels, the system can learn more fine-grained error patterns that are based on the individual error types.
\item \textbf{first language}: Previous work has experimentally demonstrated that the distribution of writing errors depends on the first language (L1) of the learner \cite{Rozovskaya2011a,Chollampatt2016a}. We investigate the usefulness of L1 as an auxiliary objective during training.

\item \textbf{part-of-speech}: POS tagging is a well-established sequence labeling task, requiring the model to disambiguate the word types based on their contexts. We use the RASP \cite{Briscoe2006} parser to automatically generate POS labels for the training data, and include them as additional objectives.
\item \textbf{grammatical relations}: We include as an auxiliary objective the type of the Grammatical Relation (GR) in which the current token is a dependent, in order to incentivise the model to learn more about semantic composition. Again we use the RASP parser, which is unlexicalised and therefore more suitable for learner data where spelling and grammatical errors are common.
\end{itemize}

\noindent Table \ref{tab:example} presents the labels for each of the auxiliary tasks for an example sentence from the FCE training data.

The auxiliary objectives introduce additional parameters into the model, in order to construct the hidden and output layers. However, these components are required only during the training process; at testing time, these can be removed and the resulting model has the same architecture and number of parameters as the baseline, with the only difference being in how the parameters were optimised.

\begin{table*}[t]
\setlength\tabcolsep{8.5pt}
\centering
\scalebox{0.9}{
\begin{tabular}{lc|cccc|cccc} \toprule
 & & \multicolumn{4}{c|}{\small{CoNLL-14 TEST1}} & \multicolumn{4}{c}{\small{CoNLL-14 TEST2}}\\
 & predicted & correct & P & R & $F_{0.5}$ & correct & P & R & $F_{0.5}$\\\midrule
R\&Y (2016) & 4449 & 683 & 15.4 & \textbf{22.8} & 16.4 & 1052 & 23.6 & \textbf{25.1} & 23.9\\\midrule
Main system & 3222 & 452 & 14.1 & 15.1 & 14.3 & 750 & 23.3 & 17.9 & 21.9\\
+ frequency & 3428 & 484 & 14.1 & 16.2 & 14.5 & 790 & 23.1 & 18.8 & 22.0\\
+ language & 3633 & 502 & 13.8 & 16.8 & 14.2 & 828 & 22.8 & 19.7 & 22.0\\
+ errors & 3582 & 557 & 15.6 & 18.6 & 16.1 & 890 & 25.0 & 21.2 & 24.0\\
+ POS & 3938 & 657 & \textbf{16.7} & 22.0 & \textbf{17.5} & 1045 & \textbf{26.5} & 24.9 & \textbf{26.2}\\
+ GR & 3945 & 593 & 15.0 & 19.8 & 15.7 & 912 & 23.2 & 21.7 & 22.8\\
+ err POS GR & 3722 & 621 & \textbf{16.7} & 20.8 & 17.4 & 979 & 26.3 & 23.3 & 25.6\\
\bottomrule
\end{tabular}
}
\caption{Error detection results on the CoNLL-14 test set using different auxiliary loss functions.}
\label{tab:resultsconll}
\end{table*}

\section{Evaluation setup and datasets}

\newcite{Rei2016} investigate a number of compositional architectures for error detection, and present state-of-the-art results using a bidirectional LSTM. 
We follow their experimental setup and investigate the impact of auxiliary loss functions on the same datasets: the First Certificate in English (FCE) dataset \cite{Yannakoudakis2011} and the CoNLL-14 shared task test set \cite{Ng2013a}.

FCE contains texts written by non-native learners of English in response to exam prompts eliciting free-text answers. The texts have been manually annotated with error types and error spans by professional examiners, which \newcite{Rei2016} convert to a binary correct/incorrect token-level labeling for error detection. 
For missing-word errors, the error label is assigned to the next word in the sequence.
The released version contains 28,731 sentences for training, 2,222 sentences for development and 2,720 sentences for testing. The development set was randomly sampled from the training data, and the test set contains texts from a different examination year.

The CoNLL-14 test set contains 50 texts annotated by two experts. Compared to FCE, the texts are more technical and are written by higher-proficiency learners.
In order to make our results comparable to \newcite{Rei2016}, we also evaluate our models on the two CoNLL-14 test annotations and train our models only on the public FCE dataset. This corresponds to their \textit{FCE-public} model that treats the CoNLL-14 dataset as an out-of-domain test set corpus.

Following the CoNLL-14 shared task, we also report $F_{0.5}$ as the main evaluation metric. 
However, while the shared task focused on correction and calculated $F_{0.5}$ over error spans using multiple annotations, we evaluate  token-level error detection performance.
Following recommendations by \newcite{Chodorow2012}, we also report the raw counts for predicted and correct tokens.

For pre-processing, all the texts are lowercased and digits are replaced with zeros for the token-level representations, although the character-based component has access to the original version of each token. 
Tokens that occur only once are mapped to a single \textit{OOV} token, which is then used to represent previously unseen tokens during testing.
The word embeddings have size 300 and are initialised with publicly available word2vec \cite{Mikolov2013a} embeddings trained on Google News. 
The LSTM hidden layers have size 200 and the task-specific hidden layers have size 50 with \textit{tanh} activation. 
The model is optimised using Adadelta \cite{Zeiler2012} and training is stopped based on the error detection $F_{0.5}$ score on the development set.
We implement the proposed framework using Theano and make the code publicly available online.\footnote{http://www.marekrei.com/projects/seqlabaux}

\section{Results}
\label{sec:results}

Table \ref{tab:resultsfce} presents the results for different system configurations trained and tested on the FCE dataset.
The first row contains results from the current state-of-the-art system by \newcite{Rei2016}, trained on the same FCE data.
The main system in our experiments is the bi-directional LSTM error detection model with character-based representations, as described in Section \ref{sec:errordet}.
We then use this model and test the effect on performance when adding each of the auxiliary loss functions described in Section \ref{sec:auxloss} to the training objective.

The auxiliary frequency loss improves performance for POS tagging \cite{Plank2016}; however in error detection the same objective does not help. While certain POS tags are more likely to belong to specific frequency classes, there is less reason to believe that word frequency provides a useful cue for error detection. 
A similar drop in performance is observed for the auxiliary loss involving the first language of the learner. It is likely that the system learns specific types of features for the L1 identification auxiliary task (such as the presence of certain words or phrases), and these are not directly useful for error detection. Investigating different architectures for incorporating the L1 as an auxiliary task is an avenue for future work.

The integration of fine-grained error types through an auxiliary loss function gives an absolute improvement of $2.1\%$ on the FCE test set. 
While the baseline only differentiates between correct and incorrect tokens, the auxiliary loss allows the system to learn feature detectors that are specialised for individual error types, thereby also making these features available to the binary error detection component.

The inclusion of POS tags and GRs gives consistent improvements over the basic configuration. Both of these tasks require the system to understand how each token behaves in the sentence, thereby encouraging it to learn higher-quality compositional representations. If the architecture is able to predict the POS tags or GR type based on context, then it can use the same features to identify irregular sequences for error detection.
The added advantage of these loss functions over the L1 and the fine-grained error types is that they can be automatically generated and require no additional manual annotation.
As far as we know, this is the first time automatically generated GR labels have been explored as objectives in a multi-task sequence labeling setting. 

Finally, we evaluate a combination system, integrating the auxiliary loss functions that performed the best on the development set. 
The combination architecture includes four different loss functions: the main binary incorrect/correct label, the fine-grained error type, the POS tag and the GR type. We left out frequency and L1, as these lowered performance on the development set.
The resulting system achieves $47.7\%$ $F_{0.5}$, which is a $4.3\%$ absolute improvement over the baseline without auxiliary loss functions, and a $6.6\%$ absolute improvement over the current state-of-the-art error detection system by \newcite{Rei2016}, trained on the same FCE dataset.

Table \ref{tab:resultsconll} contains the same set of evaluations on the two CoNLL-14 shared task annotations. 
Word frequency and L1 have nearly no effect, whereas the fine-grained error labels lead to roughly $2\%$ absolute improvement over the basic system. 
The inclusion of POS tags in the auxiliary objective consistently leads to the highest $F_{0.5}$. 
While GRs also improve performance over the main system, their overall contribution is less compared to the FCE test set, which can be explained by the different writing style in the CoNLL data. 

\begin{table}[t]
\setlength\tabcolsep{8pt}
\centering
\scalebox{0.9}{
\begin{tabular}{lccc} \toprule
 & \small{FCE} & \small{CoNLL-14} & \small{CoNLL-14}\\
Aux dataset & \small{TEST} & \small{TEST1} & \small{TEST2} \\\midrule
None & 43.4 & 14.3 & 21.9\\
CoNLL-00 & 42.5 & \textbf{15.4} & \textbf{22.3}\\
CoNLL-03 & 39.4 & 12.5 & 20.0\\
PTB-POS & \textbf{44.4} & 14.1 & 20.7\\\bottomrule
\end{tabular}}
\caption{Results on error detection when the model is pre-trained on different tasks.}
\label{tab:pretraining}
\end{table}

\section{Alternative Training Strategies}
\label{sec:altstrat}

In contrast to our approach, most previous work on multi-task learning has focused on optimising the same system on multiple datasets, each annotated with one specific type of labels. To evaluate the effectiveness of our approach, we implement two alternative multi-task learning strategies for error detection.
For these experiments, we make use of three established sequence labeling datasets that have been manually annotated for different tasks:
\begin{itemize}
\item The CoNLL 2000 dataset \cite{TjongKimSang2000} for chunking, containing sections of the Wall Street Journal and annotated with 22 different labels.
\item The CoNLL 2003 corpus \cite{TjongKimSang2003} contains texts from the Reuters Corpus and has been annotated with 8 labels for named entity recognition (NER).
\item The Penn Treebank (PTB) POS corpus \cite{Marcus1993b} contains texts from the Wall Street Journal and has been annotated with 48 POS tags.
\end{itemize}

\noindent The CoNLL-00 dataset was identified by \newcite{Bingel2017} as being the most useful additional training resource in a multi-task setting; The CoNLL-03 NER dataset has a similar label density as the error detection task; and the PTB corpus was chosen as POS tags gave consistently good performance for error detection on both the development and test sets, as demonstrated in the previous section.

\begin{table}[t]
\setlength\tabcolsep{8pt}
\centering
\scalebox{0.9}{
\begin{tabular}{lccc} \toprule
 & \small{FCE} & \small{CoNLL-14} & \small{CoNLL-14}\\
Aux dataset & \small{TEST} & \small{TEST1} & \small{TEST2} \\\midrule
None & \textbf{43.4} & \textbf{14.3} & \textbf{21.9}\\
CoNLL-00 & 30.3 & 13.0 & 17.6\\
CoNLL-03 & 31.0 & 13.1 & 18.2\\
PTB-POS & 31.9 & 11.5 & 14.9\\\bottomrule
\end{tabular}}
\caption{Results on error detection when training is alternated between the two tasks (e.g., error detection and POS tagging) and datasets.}
\label{tab:crossdomain}
\end{table}

In the first setting, each of these datasets is used to train a sequence labeling model for their respective tasks, and the resulting model is used to initialise a network for training an error detection system. While it is common to preload word embeddings from a different model, this strategy extends the idea to the compositional components of the network. Results in Table \ref{tab:pretraining} show the performance of the error detection model with and without pre-training. 
There is a slight improvement when pre-training the model on the CoNLL-00 dataset, but the increase is considerably smaller compared to the results in Section \ref{sec:results}. One of the main advantages of multi-task learning is regularisation, actively encouraging the model to learn more general-purpose features, something which is not exploited in this setting since the training happens in separate stages.

In the second set of experiments, we explore the possibility of training on the second domain and task at the same time as error detection. Similar to \newcite{Weston2008}, we randomly sample a sentence from one of the datasets and update the model parameters for that specific task. By alternating between the two tasks, the model is able to retain the regularisation benefits. However, as shown in Table \ref{tab:crossdomain}, this type of training does not improve error detection performance. One possible explanation is that the domain and writing style of these auxiliary datasets is very different from the learner writing corpus, and the model ends up optimising in an unnecessary direction. By including alternative labels on the same dataset, as in Section \ref{sec:results}, the model is able to extract more information from the domain-relevant training data and thereby achieve better results.

\section{Additional Training Data}
\label{sec:moredata}

The main benefits of multi-task learning are expected in scenarios where the available task-specific training data is limited.
However, we also investigate the effect of auxiliary objectives when training on a substantially larger training set.
More specifically, we follow \newcite{Rei2016}, who also experimented with augmenting the publicly available datasets with training data from a large proprietary corpus.
In total, we train this large model on 17.8M tokens from the Cambridge Learner Corpus (CLC, \citealt{Nicholls2003}), the NUS Corpus of Learner English (NUCLE, \citealt{Dahlmeier2013}), and the Lang-8 corpus \cite{Mizumoto2011}.

We use the same model architecture as \newcite{Rei2016}, adding only the auxiliary objective of predicting the automatically generated POS tag, which was the most successful additional objective based on the development experiments.
Table \ref{tab:moredata} contains results for evaluating this model, when trained on the large training set. 
On the FCE test data, the auxiliary objective does not provide an improvement and the model performance is comparable to the results by \newcite{Rei2016} (R\&Y).
Since most of the large training set comes from the CLC, which is quite similar to the FCE dataset, it is likely that the available training data is sufficient and the auxiliary objective does not offer an additional benefit.
However, there are considerable improvements on the CoNLL test sets, with $1.8\%$ and $1.1\%$ absolute improvements on the corresponding benchmarks.
Only small amounts of the training data are similar to the CoNLL dataset, and including the auxiliary objective has provided a more robust model that delivers better performance on different writing styles.

\begin{table}[t]
\setlength\tabcolsep{7pt}
\centering
\scalebox{0.9}{
\begin{tabular}{l|c|ccc} \toprule
 & \small{R\&Y} $F_{0.5}$ & P & R & $F_{0.5}$ \\\midrule
\small{FCE DEV} & 60.7 & 75.1 & 35.1 & \textbf{61.2}\\
\small{FCE TEST} & \textbf{64.3} & 78.4 & 37.0 & 64.1\\
\small{CoNLL TEST1} & 34.3 & 44.7 & 20.5 & \textbf{36.1}\\
\small{CoNLL TEST2} & 44.0 & 63.8 & 20.8 & \textbf{45.1}\\\bottomrule
\end{tabular}}
\caption{Error detection results using auxiliary objectives, trained on additional data.}
\label{tab:moredata}
\end{table}

\section{Previous Work}

\noindent{\bf Error detection:} Early error detection systems were based on manually constructed error grammars and mal-rules (e.g., \citealt{foster2004parsing}). More recent approaches have exploited error-annotated learner corpora and primarily treated the task as a classification problem over vectors of contextual, lexical and syntactic features extracted from a fixed window around the target token. Most work has focused on error-type specific detection models, and in particular on models detecting preposition and article errors, which are among the most frequent ones in non-native English learner writing \cite{chodorow2007,de2008classifier,han2010using,tetreault2010using,han2006, tetreault2008ups,gamon2008using,gamon2010using,kochmar2014detecting,leacock2014automated}. 
Maximum entropy models along with rule-based filters account for a substantial proportion of utilized techniques. Error detection models have also been an integral component of essay scoring systems and writing instruction tools \cite{burstein2004automated,andersen2013developing,attali2006automated}.

The Helping Our Own (HOO) 2011 shared task on error detection and correction focused on a set of different errors \cite{dale2011helping}, though most systems were type specific and targeted closed-class errors. In the following year, the HOO 2012 shared task only focused on correcting preposition and determiner errors \cite{dale2012hoo}. 
The recent CoNLL shared tasks \cite{ng2013conll,Ng2014conll} focused on error correction rather than detection: CoNLL-13 targeted correcting noun number, verb form and subject-verb agreement errors, in addition to preposition and determiner errors made by non-native learners of English, whereas CoNLL-14 expanded to correction of all errors regardless of type. 
Core components of the top two systems across the CoNLL correction shared tasks include Average Perceptrons, L1 error correction priors in Naive Bayes models, and joint inference capturing interactions between errors (e.g., noun number and verb agreement errors) \cite{rozovskaya2014illinois}, as well as phrase-based statistical machine translation, under the hypothesis that incorrect source sentences can be ``translated'' to correct target sentences \cite{felice2014,grundkiewicz2014amu}.

The work that is most closely related to our own is the one by \newcite{Rei2016}, who investigate a number of compositional architectures for error detection, and propose a framework based on bidirectional LSTMs. 
In this work, we used their system architecture as a baseline, compared our model to their results in Sections \ref{sec:results} and \ref{sec:moredata}, and showed that multi-task learning can further improve performance and allow the model to generalise better.

\noindent{\bf Multi-task learning:} Multi-task learning was first proposed by \newcite{Caruana1998} and has since been applied to many language processing tasks and neural network architectures. 
For example, \newcite{Weston2008} constructed a convolutional architecture that shared some weights between tasks such as POS tagging, NER and chunking. 
Whereas their model only shared word embeddings, our approach focuses on learning better compositional features through a shared bidirectional LSTM.

\newcite{Luong2016} explored a multi-task architecture for sequence-to-sequence learning where encoders and decoders in different languages are trained jointly using the same semantic representation space. 
\newcite{Klerke2016} used eye tracking measurements as a secondary task in order to improve a model for sentence compression.
\newcite{Bingel2017} explored beneficial task relationships for training multitask models on different datasets. 
All of these architectures are trained by randomly switching between different tasks and updating parameters based on the corresponding dataset. In contrast, we treat alternative tasks as auxiliary objectives on the same dataset, which is beneficial for error detection  (Section \ref{sec:altstrat}).

There has been some research on using auxiliary training objectives in the context of other tasks. 
\newcite{Cheng2015a} described a system for detecting out-of-vocabulary names by also predicting the next word in the sequence. 
\newcite{Plank2016} predicted the frequency of each word together with the POS, and showed that this can improve tagging accuracy on low-frequency words.
However, we are the first to explore the auxiliary objectives described in Section \ref{sec:auxloss} in the context of error detection.

\section{Conclusion}

We have described a method for integrating auxiliary loss functions with a neural sequence labeling framework, in order to improve error detection in learner writing. 
While predicting binary error labels, the model also learns to predict additional linguistic information for each token, allowing it to discover compositional features that can be exploited for error detection.
We performed a systematic comparison of possible auxiliary labels, which are either available in existing annotations or can be generated automatically. 
Our experiments showed that POS tags, grammatical relations and error types gave the largest benefit for error detection, and combining them together improved the results further.

We compared this training method to two other multi-task approaches: learning sequence labeling models on related tasks and using them to initialise the error detection model; and training on multiple tasks and datasets by randomly switching between them. Both of these methods were outperformed by our proposed approach using auxiliary labels on the same dataset -- the latter has the benefit of regularising the model with a different task, while also keeping the training data in-domain.

While the main benefits of multi-task learning are expected in scenarios where the available task-specific training data is limited, we found that error detection benefits from additional labels even with large training sets.
Successful error detection systems have to learn about language composition, and introducing an additional objective encourages the model to train more general composition functions and better word representations.
The error detection model, which also learns to predict automatically generated POS tags, achieved improved performance on both CoNLL-14 benchmarks.
A useful direction for future work would be to investigate 
dynamic weighting strategies for auxiliary objectives that allow the network to initially benefit from various available labels, and then specialise to performing the main task.

\bibliography{references}
\bibliographystyle{emnlp_natbib}

\end{document}